\documentclass[11pt,a4paper]{article}
\usepackage[hyperref]{acl2021}
\usepackage{times}
\usepackage{latexsym}
\usepackage{graphicx}
\usepackage{caption}
\usepackage{subcaption}
\usepackage{booktabs}
\usepackage{multirow}
\usepackage{multicol}
\usepackage{amsmath}
\usepackage{courier}
\usepackage{url}
\usepackage{CJKutf8}

\usepackage{microtype}

\usepackage{enumitem}

\usepackage{tikz}
\usepackage{pgfplots}
\pgfplotsset{width=7.5cm,compat=1.12}
\usepgfplotslibrary{fillbetween}

\usepackage{makecell}

\usepackage{xcolor}

\aclfinalcopy 

\title{Two Approaches to Building Collaborative, Task-Oriented Dialog Agents through Self-Play}

\author{Arkady Arkhangorodsky, Scot Fang, Victoria Knight, \\
\bf Ajay Nagesh, Maria Ryskina, Kevin Knight  \\
DiDi Labs\\
4640 Admiralty Way \\
Marina del Rey, CA 90292}

\date{}

\begin{document}

\maketitle

\begin{abstract}
Task-oriented dialog systems are often trained on human/human dialogs, such as collected from Wizard-of-Oz interfaces.  
However, human/human corpora are frequently too small for supervised training to be effective.  This paper investigates two approaches to training agent-bots and user-bots through self-play, in which they autonomously explore an API environment, discovering communication strategies that enable them to solve the task. We give empirical results for both reinforcement learning and game-theoretic equilibrium finding.
\end{abstract}

\section{Introduction}

In this paper, we study dialog self-play, in which an agent-bot and user-bot develop communication and action strategies to solve task-oriented dialog problems.  

We adopt a highly-simplified version of the trip-booking dialog domain of \newcite{meep}, where the user (passenger) talks about their desired destination (``I want to go to Starbucks on Venice Boulevard''), and the agent (autonomous vehicle or personal assistant) uses Google Map APIs and utterances to propose and confirm destinations.  At the end of the dialog, we check whether the booked destination is correct, from the user's point of view. A sample human/human dialog from this domain is shown in Figure~\ref{dialogs}.  Only the verbal actions are shown.  Not shown are the Google Map API calls that the agent uses, or what those calls return.  The example highlights several challenging features of this task domain: millions of places, ambiguity of place names, calculation of distances, need for confirmation, etc.

\begin{figure}
\begin{scriptsize}
    \begin{verbatim}
User:   I want to go to Starbucks on Venice 
        Boulevard.
Agent:  There is a Starbucks in Mar Vista.
        Are you okay with that one?
User:   Is it the one across from Coffee Connection?
Agent:  Starbucks is 141 feet away from Coffee 
        Connection.
        It will take us 10 minutes to get there.
        Shall we go?
User:   Great, thanks.
Agent:  Great, we are going to Starbucks.
    \end{verbatim}
\end{scriptsize}
\vspace*{-0.2in}
\caption{Sample human/human dialog in the trip-booking domain \cite{meep}.  A user (passenger) works with an agent (autonomous vehicle or personal assistant) to book a trip destination using a speech interface.}
\label{dialogs}
\end{figure}


Given a human/human corpus, we can use machine learning to build an automatic agent that imitates a human agent. However, hundreds or thousands of human/human dialogs are often insufficient to train high-quality agents.  This kind of training gap is commonly addressed with a user simulator \cite{dialog7,dialog3,dialog67,dialog55,dialog79,dialog36}.  An automatic agent can interact with a user simulator to effectively generate many more positive examples of tasks completed.  Reinforcement learning (RL) is commonly used \cite{sutton1998introduction}.

Indeed, in domains we are interested in, we can hand-build a user simulator.  But in that case,  we find that we might as well instead just hand-build an agent simulator, and dispense with training.  We can also train a user simulator on human/human dialogs, but we again run into data sparsity.

In this paper, we address data sparsity and limit handcrafting by investigating self-play, in which an agent-bot and a user-bot learn strategies from scratch.
We supply randomly-generated destinations to the user-bot and reward both bots if (after the dialog) the agent-bot drives the user-bot to the correct place.  We provide the user-bot with a push-button environment similar to the agent side.


In self-play for task-oriented dialog, we face three main challenges:

\begin{itemize}
\item Designing a learning scheme that allows the bots to obtain {\em full rewards}, by autonomously solving the tasks given to them.
\item Preventing the two bots from developing a {\em secret language}. After they are trained, we want the bots to be able to interact correctly with normal human users.
\item Preventing the two bots from developing a {\em single, narrow solution}.  The learned bots should be able to react correctly to a wide range of dialog situations.
\end{itemize}

\section{Related Work and Contributions}

Inspirational examples of self-play include competitive games like backgammon \cite{10.1145/203330.203343}, Go \cite{silver2017mastering}, and poker \cite{brown2018superhuman}, and collaborative games like Hanabi \cite{BARD2020103216}.

In collaborative, task-oriented dialog, self-play has been used primarily to create synthetic data \cite{shah2018building,shah-etal-2018-bootstrapping,majumdar}. In addition, it has been used to train agents in particular types of structured dialogs, such as recommendation \cite{kang-etal-2019-recommendation} and negotiation \cite{lewis-etal-2017-deal,Jang_Lee_Kim_2020}.  Further, \newcite{liu-lane-2018-adversarial} use adversarial learning to supplement sparse reward signals.  The closest work to ours is Section~6.2 of \newcite{dialog8}, which briefly describes a self-play formulation in an air-travel domain with a vast human/human dialog corpus.

The contributions of the present paper are:

\begin{itemize}
\item We implement fully-autonomous self-play in a collaborative, task-oriented dialog domain, addressing the challenges above.
\item We compare two methods, one based on reinforcement learning methods, and one based on game-theoretic equilibrium-finding.
\end{itemize}

We do not address the entire trip-booking domain in this paper.  Rather, we use a highly-simplified form of the problem to closely study the behavior of learning algorithms.


\section{Simplified Trip-Booking}

Our domain is characterized as follows.

\begin{itemize}
\item Desired destination: Randomly selected as either {\em Starbucks} or {\em Peet's}.
\item The user-bot (passenger) has two verbal actions: {\em Say-Starbucks} and {\em Say-Peet's}.
\item The agent-bot (driver) has two API calls: {\em Drive-Starbucks} and {\em Drive-Peet's}.  In this domain, the agent-bot does not talk.
\item Reward: Both bots get -1.0 if agent-bot drives to the wrong destination, +1.1 if the agent-bot drives to the right destination (and the user-bot mentioned that destination), +1.0 if the agent-bot drives to the right destination (despite the user-bot saying the wrong destination).
\item Interaction: The user-bot performs a single action, the agent-bot performs a single action, and rewards are then assigned.
\end{itemize}

Note that the agent-bot can only partially observe the environment.  It cannot see the desired destination.  It is the user-bot's job to relay that information to the agent-bot. We supply a stream of desired destinations to the user-bot, and we want the two bots to converge on strategies that yield good rewards.  


{\bf RL solution}. We first attack this problem with RL. Initially, the interactions between the bots are nonsensical and low-reward.  For example, Destination: {\em Starbucks}, User: {\em Say-Starbucks}, Agent: {\em Drive-Peet's}, Reward: -1.0.

To improve the bots' policies, we use the PG (policy gradient) and PPO (promixal policy optimization) algorithms as implemented in {\em rllib}~\cite{rllib} with default parameters. We run 90 iterations of the training loop with a batch size of 500 actions, resulting in 27,000 dialogs in a random restart. On top of that, we execute 100 restarts for each algorithm.   

\begin{figure*}[h]
\centering
\begin{tabular}{|l|r|r|r|} \hline 
Method &  restarts w/  & restarts w/ & avg time to converge \\ 
 &  +1.1 reward & +1.0 reward & (seconds) \\ \hline
PG (policy gradient) & 97/100 & 3/100 & 63.4 sec \\ \hline
PPO (promixal policy optimization) & 84/100 & 16/100 & 81.6 sec \\ \hline
Game-theoretic equilibrium finding & 1/1 & 0/1 & $<$ 1 sec\\ \hline
\end{tabular}    
\caption{Self-play results. For each restart in the RL experiments (first two rows), we consider it to achieve a particular reward if at least 2700 of the last 3000 episodes of training obtain that reward. Time to converge for each restart is the first time the reward, averaged over the last 300 episodes, exceeds +1.095; or, complete training time if the restart does not achieve the full reward.} 
\label{demo-results}
\end{figure*}

\begin{figure*}
\centering
\includegraphics[scale=1.2]{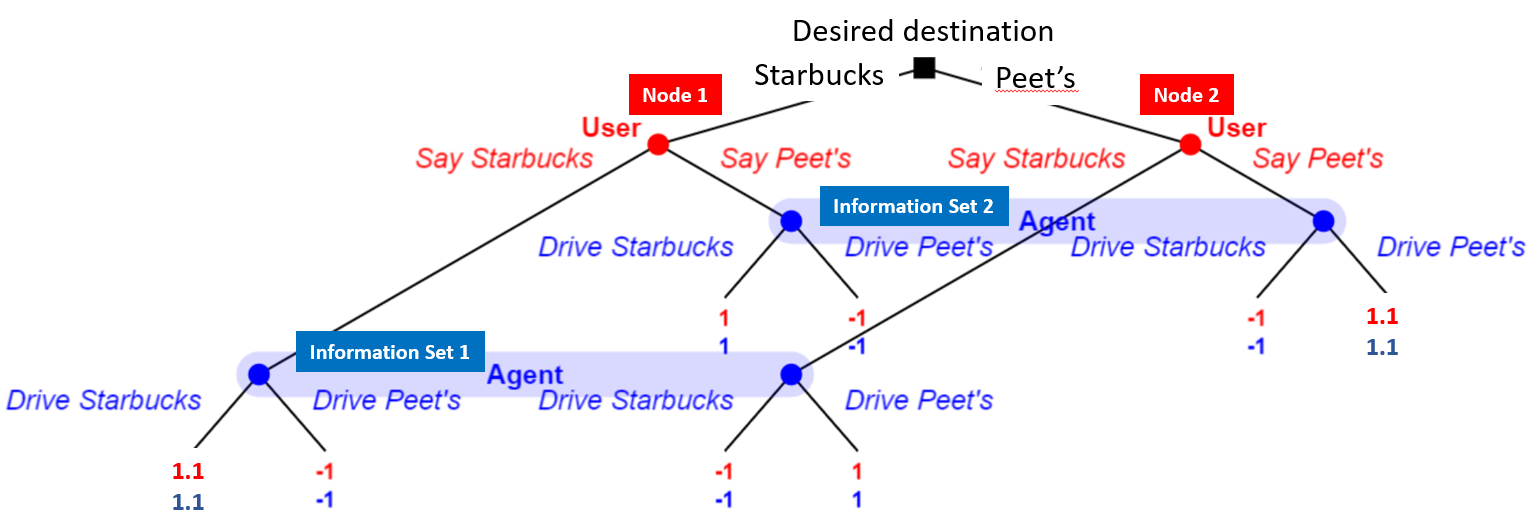}
\caption{Extensive game tree. There is one chance node (black), 2 user decision nodes (red), and 4 agent decision nodes grouped into 2 information sets (blue). The tree has 8 leaf nodes which store rewards.}
\label{extensive-tree}
\end{figure*}

\begin{figure*}
\centering
\begin{footnotesize}
\begin{tabular}{|l|c|c|c|c|c|c|c|c|l|} \hline
\multirow{2}{*}{Equilibrium} & \multicolumn{2}{c|}{User Node 1} & \multicolumn{2}{c|}{User Node 2} & \multicolumn{2}{c|}{Agent Info Set 1} &\multicolumn{2}{c|}{Agent Info Set 2} & \multirow{2}{*}{Expected Reward} \\ 
& Say-S & Say-P & Say-S & Say-P & Drive-S & Drive-P & Drive-S & Drive-P &   \\ \hline
\#1 & 1 & 0 & 0 & 1 & 1 & 0 & 0 & 1 & +1.1 \\ \hline
\#2 & 0 & 1 & 1 & 0 & 0 & 1 & 1 & 0 & +1.0 \\ \hline
\#3 & 1 & 0 & 1 & 0 & 1 & 0 & 1 & 0 & +0.05 \\ \hline
\#4 & 0 & 1 & 0 & 1 & 0 & 1 & 0 & 1 & +0.05 \\ \hline
\#5 & 20/21 & 1/21 & 1 & 0 & 20/21 & 1/21 & 1 & 0 & +0.0476 \\ \hline
\#6 & 0 & 1 & 1/21 & 20/21 & 0 & 1 & 1/21 & 20/21 & +0.0476 \\ \hline
\#7 & 20/41 & 21/41 & 21/41 & 20/41  & 20/41 & 21/41 & 21/41 & 20/41 & +0.0249 \\ \hline
\end{tabular}
\end{footnotesize}
\caption{Equilibria computed for the game tree in Figure~\ref{extensive-tree}, giving strategies for User and Agent.  As an example, in Equilibrium \#5, the user at Node 1 flips a coin, electing {\em Say-Starbucks} with probability 20/21 and {\em Say-Peet's} with probability 1/21.  Likewise for Equilibrium \#5, an agent in Information Set 2 (meaning ``user said Peet's, but actual destination is unknown'') will always elect {\em Drive-Starbucks}.}
\label{equilibria}
\end{figure*}

\begin{figure}[h]
\centering
\includegraphics[scale=1.0]{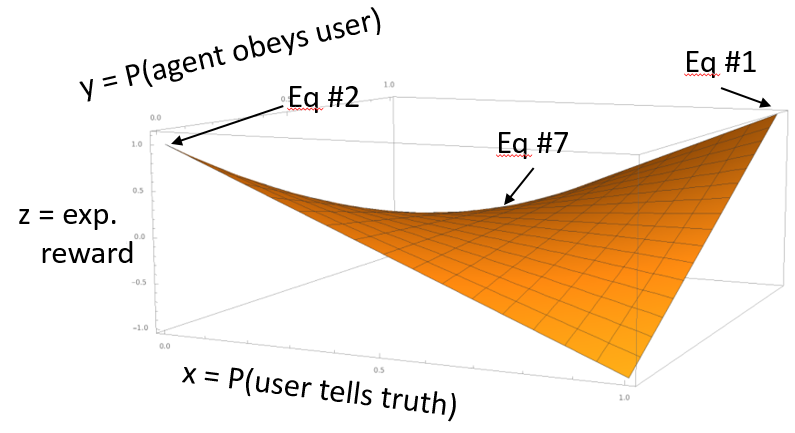}
\caption{Three visualized equilibria.  The x-axis gives mixed strategies for the user (1.0 = always faithfully relate desired destination, 0.0 always lie about destination).  The y-axis gives mixed strategies for the agent (1.0 = always obey user, 0.0 always disobey user).}
\label{eq-viz}
\end{figure}

Figure~\ref{demo-results} shows results. We note two things:

\begin{itemize}
\item Some proportion of random restarts result in bots that are unable to obtain the full +1.1 reward. 
\item The simpler PG algorithm achieves +1.1 reward in more restarts than the PPO algorithm.
\end{itemize}

Bots that achieve a suboptimal +1.0 reward do so by developing a secret language.  When the destination is {\em Starbucks}, the user says {\em Peet's}, yet the agent drives to {\em Starbucks}.  We call this ``opposite day.''  We might expect the agents to switch to strategies that deliver the full reward (+1.1) if we manipulate exploration parameters, but in practice, we find this difficult.


{\bf Observation}. We note that neither dialog bot explicitly models the possible strategies of the other.  Each bot simply tries to cope, during training, with an environment that appears to respond unpredictably and non-deterministically.  In fact, the other bot is unwittingly causing this apparent mayhem as it updates its own policy.

{\bf Game theory solution}. We use game theory to explicate these strategic interactions.  Theoretical frameworks for the use of game theory in dialog include \newcite{game-theoretic-account}, \newcite{lewin2000formal}, and \newcite{5940726}, while \newcite{barlier2015human} carry out experiments in adversarial-type dialogs.  Here, we apply game theory to collaborative, task-oriented dialog, where an agent works with a user to solve the user's problem.

Figure~\ref{extensive-tree} shows an extensive tree for the domain.  The first node (black) is a chance node that selects the desired destination.  At the next level, the user (red) selects a say-action.  Finally, the agent (blue) selects a drive-action.  Leaf nodes contain the relevant rewards.

The agent has imperfect information, because it cannot see the desired destination.  It can only learn about it from the user.  Thus, the agent cannot distinguish between certain nodes, and it must apply the same strategy in those cases.  We model this with standard {\em information sets} (blue shaded ovals).  At the two lowest nodes, the agent only knows that the user said ``Starbucks,'' so it cannot plan to go left at one node and right at the other.

Like an RL policy, a strategy is a comprehensive, probabilistic map from environment to action.  With games of perfect information, optimal strategies for both players can be obtained by simply backing up the rewards through the tree (backward induction or minimax search).  With imperfect information, the job requires a more complex algorithm \cite{avis}.

We first compute equilibria.  An equilibrium is a pair of strategies where neither player can unilaterally improve their reward.  There are seven such equilibria, shown in Figure~\ref{equilibria}, which we compute with the \texttt{gambit-enummixed} method in the open-source Gambit software \cite{gambit}.

Equilibrium \#1 represents the pair of strategies that obtain +1.1 reward (the user truthfully relates the destination, and the agent believes it).  Equilibrium \#2 represents opposite day, with +1.0 reward.  Equilibrium \#3 is ``always say Starbucks, always drive to Starbucks.''  The agent cannot improve things unilaterally, either by driving to Peet's or flipping a coin.  Likewise, the user cannot improve things unilaterally, as long as the agent is always driving to Starbucks.

Equilibrium \#7 involves mixed strategies on both sides.  The user mostly lies about the destination, and the agent mostly disobeys the user.  Perhaps surprisingly, neither agent can improve things unilaterally, even though a small concerted change (by both) produces a higher reward and leads to a non-equilibrium.  

Figure~\ref{eq-viz} illustrates that Equilbrium \#7 is a saddle-point. To make a 3-dimensional figure, we reduce the bots' options. Here, the x-axis records the user's probability of telling the truth, which excludes strategies like ``always say Starbucks'' (see \#3) from this plot.

In the game-theoretic framework, ``self-play'' is less play and more equation solving:

\begin{enumerate}
\item Enumerate all equilibria. 
\item Choose the highest-reward equilibrium.
\item Record the user's and agent's strategies.
\item Apply those strategies on randomly-generated destinations.
\end{enumerate}

\noindent
Figure~\ref{demo-results} shows this method is effective, requiring less than one second to compute equilibria.

\section{Conclusions and Future Work}

We present results on self-play for task-oriented dialog with a highly-simplified form of trip-booking:

\begin{itemize}
\item The bots are able to obtain full rewards autonomously, without human-produced training data. We give empirical results for both RL and game-theoretic equilibrium finding.
\item We address secret languages by rewarding the user (passenger) for telling the truth, augmenting the standard reward for task completion.
\item The game-theoretic solution not only locates the global maximum reward, it also avoids narrow solutions by computing a pair of complete strategies. 
\end{itemize}


In future work, we would like to scale these approaches to self-play in complex, collaborative, task-oriented domains such as restaurant booking and full-scale trip booking. In such domains, extensive game trees will be quite large and likely require algorithms such as those used to solve complex games like poker \cite{brown2018superhuman}.

\bibliography{self-play}
\bibliographystyle{acl_natbib}

\end{document}